\documentclass[runningheads]{llncs}
\usepackage{amssymb}
\usepackage{hyperref}
\usepackage[T1]{fontenc}
\usepackage{graphicx,verbatim}
\begin{document}
\title{EAD: An EEG Adapter for Automated Classification}
%

\author{
    Pushapdeep Singh\inst{1} \and
    Jyoti Nigam\inst{1} \and
    Medicherla Vamsi Krishna\inst{2} \and
    Arnav Bhavsar\inst{1} \and
    Aditya Nigam\inst{1}
}
\authorrunning{P. Singh et al.}

\institute{
    School of Computing and Electrical Engineering, IIT Mandi, Mandi, India \\
    \url{http://www.iitmandi.ac.in} \\
    \email{erpd2201@students.iitmandi.ac.in, jyoti\_nigam@projects.iitmandi.ac.in, arnav@iitmandi.ac.in, aditya@iitmandi.ac.in}
    \and 
     Swiss RE GBS India Pvt. Ltd., Hyderabad, India \\ 
    \email{m.k.vamsi@gmail.com} 
}
    
\maketitle              
\begin{abstract}

While electroencephalography (EEG) has been a popular modality for neural decoding, it often involves task specific acquisition of the EEG data. This poses challenges for the development of a unified pipeline to learn embeddings for various EEG signal classification, which is often involved in various decoding tasks. Traditionally, EEG classification involves the step of signal preprocessing and the use of deep learning techniques, which are highly dependent on the number of EEG channels in each sample. However, the same pipeline cannot be applied even if the EEG data is collected for the same experiment but with different acquisition devices. This necessitates the development of a framework for learning EEG embeddings, which could be highly beneficial for tasks involving multiple EEG samples for the same task but with varying numbers of EEG channels.In this work, we propose EEG Adapter (EAD), a flexible framework compatible with any signal acquisition device. More specifically, we leverage a recent EEG foundational model with significant adaptations to learn robust representations from the EEG data for the classification task. We evaluate EAD on two publicly available datasets achieving state-of-the-art accuracies 99.33\% and 92.31\% on EEG-ImageNet and BrainLat respectively. This illustrates the effectiveness of the proposed framework across diverse EEG datasets containing two different perception tasks: stimulus and resting-state EEG signals. We also perform zero-shot EEG classification on EEG-ImageNet task to demonstrate the generalization capability of the proposed approach. 


\end{abstract}

\section{Introduction}
The field of Brain-Computer Interfaces (BCIs) has experienced a significant increase in interest and research. This rapidly advancing domain promises to revolutionize human-computer interaction by leveraging our expanding knowledge of brain activity. Among the various methods for recording neural signals, Electroencephalography (EEG) has garnered considerable attention for its non-invasive capability to capture detailed spatio-temporal neural signaling data.


EEG has shown considerable promise in diverse applications such as seizure epilepsy classification \cite{Boonyakitanont}, motor imagery recognition \cite{Amin}, emotion analysis \cite{Suhaimi}, and auditory attention detection \cite{Biesmans}. Each of these applications necessitates specialized acquisition and modeling, as a universal approach is not feasible.

\begin{figure}
\centering
\includegraphics[width=0.6\textwidth]{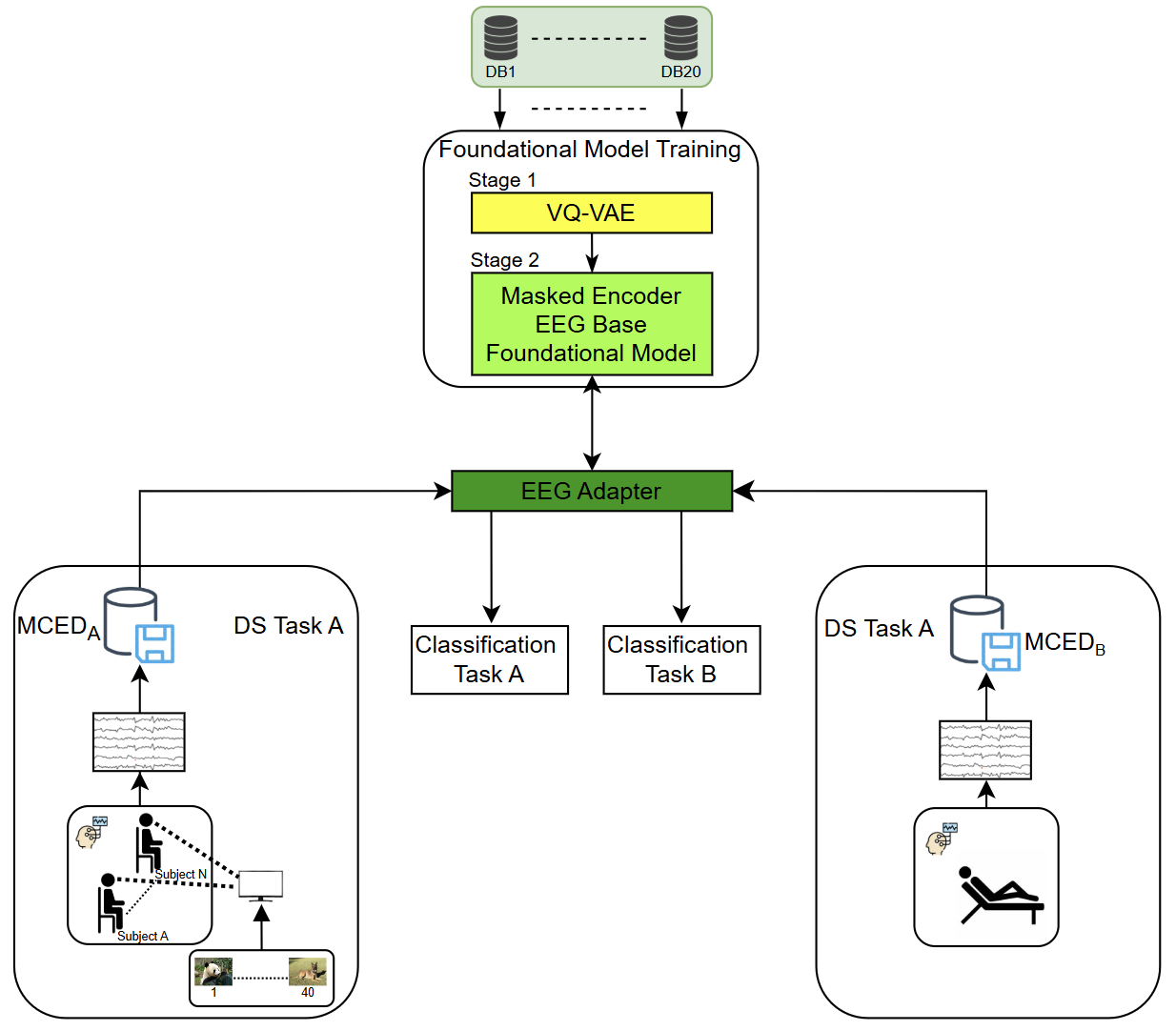}
\caption{The figure demonstrates our approach to utilize the EAD for classification tasks. We utilize two setups resting state and image decoding multichannel EEG dataset (MCED) to showcase the effectiveness of our approach.} \label{fig1}
\end{figure}

Our framework towards generalization is based on a recent foundational model, LaBraM \cite{jiang2024large}, designed for EEG feature extraction, to learn robust embeddings from raw EEG signals. However, although these embeddings contain rich neural information, they are not inherently optimized for specific downstream tasks. To address this, we introduce an EEG adapter framework that refines and transforms the LaBram-generated embeddings, making them more suitable for targeted applications such as classifying stimulus, resting-state or image retrieval from EEG. These representations can be learnt for EEG classification tasks in diverse domains such as image perception, clinical diagnosis etc. By integrating an adapter network, we enhance the adaptability and usability of EEG embeddings, facilitating more effective EEG-based visual decoding and downstream processing. Figure \ref{fig1} presents the overview of our approach to utilize the EAD for classification tasks.


To summarize our contributions include:
\begin{enumerate} 
    \item A unified framework to process diverse EEG datasets for two distinct perception tasks: stimulus-based EEG signals and resting-state EEG signals.
    \item Demonstrating the adaptability of the representations obtained from our EEG feature extraction framework, we present results on classification performance on two publicly available datasets, outperforming existing state-of-the-art methods on EEG-ImageNet \cite{Spampinato2017} and BrainLat \cite{Prado2023}, respectively. 
    \item We implement a rigorous evaluation by training our model for EEG-ImageNet on 34 classes out of 40 while reserving 6 unseen classes for testing on visual decoding tasks and subject-independent classification of disease on resting-state EEG tasks. This setup ensures a robust assessment of the proposed adapter, showcasing its generalizability. \end{enumerate}

\section{Related Work}
Brain visual decoding task involves collecting data from a subject while viewing different types of images on a screen. In the initial work, \cite{Spampinato2017,Palazzo2021} introduced the task of perceptual brain decoding on the EEG-ImageNet dataset using deep learning models for EEG classification corresponding to image perception. Building on this, \cite{Kavasidis2017,Palazzo2017} utilized generative models learnt over such embeddings for image synthesis, which garnered significant attention from the research community in visual brain decoding tasks. The crucial aspect of this task is to first learn the EEG representations, which will eventually be used as a condition for various image generation techniques \cite{Mishra2023,Singh2024}. 

Apart from visual brain decoding tasks, EEG has numerous applications in the medical domain, such as seizure detection \cite{Strein2019} and dementia diagnosis \cite{Cassani2018}.  For clinical diagnosis data is often  collected at resting state.

Previous works have focused on developing new algorithms for very specific tasks, limiting their applicability to other tasks. This narrow focus often results in solutions that cannot be easily adapted to different contexts or datasets. To address this limitation, we propose creating an adapter on top of the EEG foundational model, inspired by the success of adapters in the image generation domain \cite{Mou2024}. This approach aims to enhance the versatility and generalizability of the model, making it applicable to a wider range of tasks.

\section{Methodology}
In this section, we describe our proposed framework for adapting EEG foundational models which enables them to perform the downstream tasks.

\subsection{Analysis of LaBraM}
The LaBraM model \cite{jiang2024large} is an efficient approach for learning generic EEG signal representations. It utilizes codebook representations learned during VQ-VAE training \cite{VanDenOord2017} , where the VQ-VAE reconstructs the EEG signal's frequency domain during the training process. 


Trained VQ-VAE's encoder is utilized as a supervisor for training Masked Encoder comprising of temporal convolution in tandem with transformer referred to as the Base Foundation Model (BFM). For subsequent downstream classification tasks, fine-tuning of the BFM can be done for task specific multichannel EEG dataset.
Our approach focuses on analyzing channel alignment and automating the distillation of the EEG input data. 

In the paper \cite{jiang2024large}, the authors propose a method to utilize upto $128$ raw channels by modifying the positional embedding. We also present the results of this method alongside our proposed approach. EAD minimizes the preprocessing required for learning a discriminative embedding on EEG signals, leading to an end-to-end framework that can be integrated with both low-density and high-density EEG signal capturing devices. The conducted experiments demonstrate the results for manual selection, modified BFM, and automated distillation. 

\subsection{Manual Channel Alignment}
The acquisition configuration for different tasks and datasets often differ. Thus, we need to have a method for channel alignment to use the same model. The first method we employed was to manually select the channels that are close to the montage configuration proposed by the BFM as shown in Table \ref{table:montages}. We take the raw signals from the EEG data and match the montage electrode locations of the dataset with the anatomically closest montage electrode locations of the evaluation dataset TUAB \cite{Obeid2016} used in \cite{jiang2024large}. 

From this point, we employ two methods: 
\begin{enumerate}
    \item The first method involves identifying an approximate match within the provided montage using the anatomical placement of the electrodes. To match the input shape  of the BFM, we either trim or repeat the signal per channel to match the signal length.
    \item The second approach combines the signals per channel from the four surrounding electrodes into a single composite signal for further analysis. For mixing channels we trim or repeat each signal per channel equally and concatenate on time axis.
\end{enumerate}

\begin{table}[h!]
\centering
\caption{Mapping of EEG-ImageNet montage for Manual channel selection including mixing with BFM's TUAB montage.}\label{table:montages}
\begin{tabular}{|l|l|l|}
\hline
\textbf{BFM} & \textbf{Manual Channel Alignment} & \textbf{Manual Channel Alignment Mixing} \\ \hline
FP1 & Fp1 & Fp1, Afp1, AF3, AF7, AFF5h \\ \hline
FP2 & Fp2 & Fp2, Afp2, AF4, AF8, AFF6h \\ \hline
F3 & F3 & F3, F5, F1, FFC5h, FFC3h \\ \hline
F4 & F4 & F4, F2, F6, FFC4h, FFC6h \\ \hline
C3 & C3 & C3, C5, C1, CCP5h, CCP3h \\ \hline
C4 & C4 & C4, C6, C2, CCP4h, CCP6h \\ \hline
P3 & P3 & P3, P1, P5, CPP5h, CPP3h \\ \hline
P4 & P4 & P4, P2, P6, CPP4h, CPP6h \\ \hline
O1 & O1 & O1, POO1, PO3, PO7, POO9h \\ \hline
O2 & O2 & O2, POO2, PO4, PO8, POO10h \\ \hline
F7 & F7 & F7, F5, F9, FFT9h, FFT7h \\ \hline
F8 & F8 & F8, F6, F10, FFT8h, FFT10h \\ \hline
T3 & T7 & T7, TTP7h, C5, FTT7h, FTT9h \\ \hline
T4 & T8 & T8, TTP8h, C6, FTT8h, FTT10h \\ \hline
T5 & TP7 & TP7, TTP7h, CP5, TPP7h, TPP9h \\ \hline
T6 & TP8 & TP8, TTP8h, CP6, TPP8h, TPP10h \\ \hline
A1 & TP9 & TP9, TP7, T7, FTT9h, FT9 \\ \hline
A2 & TP10 & TP10, TP8, T8, FTT10h, FT10 \\ \hline
FZ & Fz & Fz, AFF1h, AFF2h, FFC1h, FFC2h \\ \hline
CZ & Cz & Cz, FCC1h, FCC2h, CCP1h, CCP2h \\ \hline
PZ & Pz & Pz, CPP1h, CPP2h, POO1h, PPO2h \\ \hline
T1 & TTP7h & TTP7h, C5, TP7, CP5, CCP5h \\ \hline
T2 & TTP8h & TTP8h, C6, TP8, CP6, CCP6h \\ \hline
\end{tabular}
\end{table}


We fine-tune the pre-trained BFM to learn the representation of EEG data from the manually aligned channels. With the mixing approach, we achieve better accuracy on the EEG-ImageNet dataset. However, even if we only take the approximate montage, the results are still very encouraging.


\begin{figure}
\centering
\includegraphics[width=0.6\textwidth]{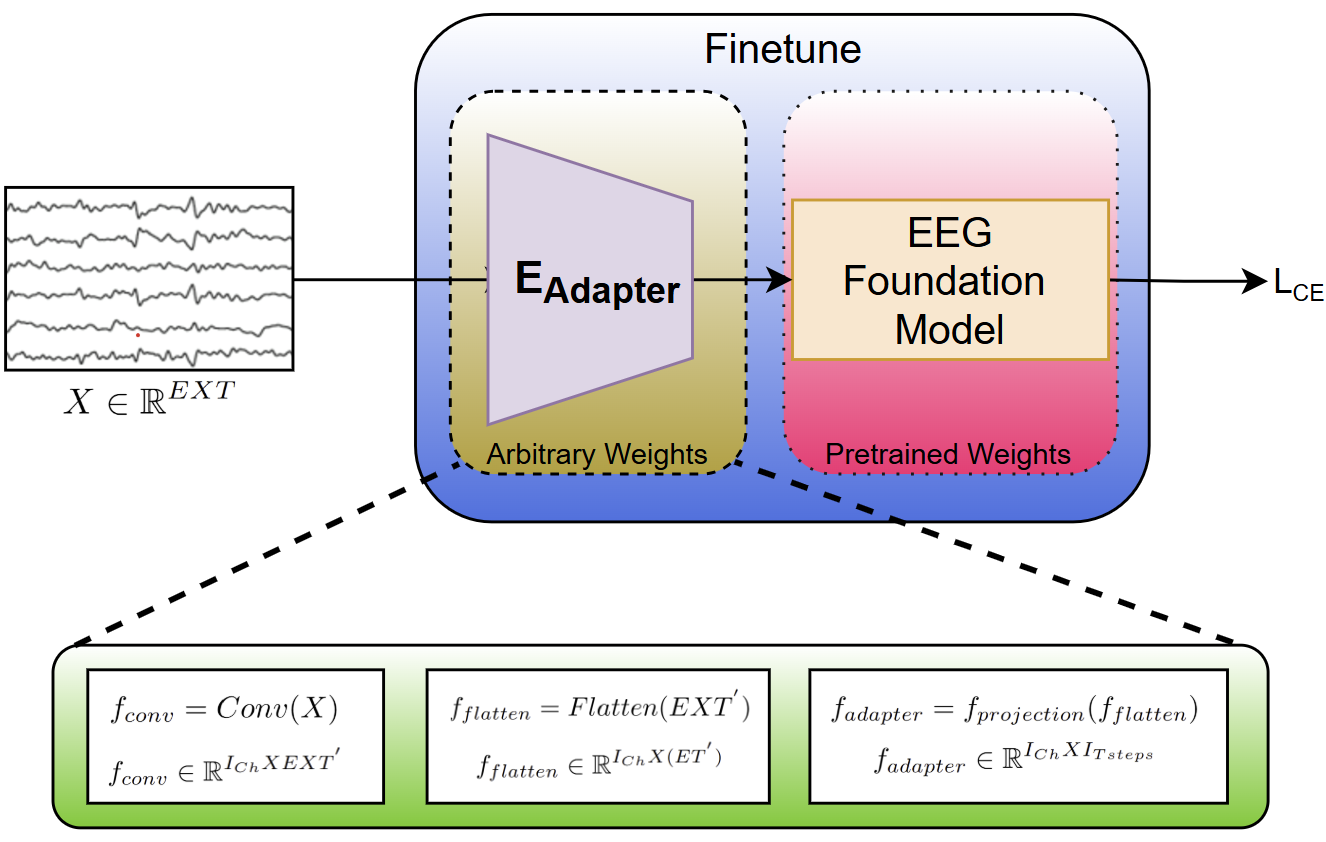}
\caption{This image portrays a magnified view of the EAD architecture. $X$ refers to the input EEG signal with $E$ channels and $T$ time steps. $T^{'}$ is dependent upon the convolution kernel, $I_{Ch}$ is the number of input channels and $I_{Tsteps}$ is the number of time steps. $L_{CE}$ is categorical cross entropy loss.} \label{fig2}
\end{figure}

\subsection{Automatic Channel Distillation using EAD}
Since we achieved the best results on the manually aligned composite signal, we hypothesize that this is due to the additional information present in the composite signals. This realization provides us with a direction towards automatic distillation.
Thus, to enhance the manual channel mixing method, we propose an EAD to automatically distill EEG signals into the BFM. This adapter can be used in conjunction with a BFM to create an end-to-end automated pipeline for EEG classification tasks as shown in Figure \ref{fig2}. 

To achieve this, consider an EEG sensor with $E$ channels and $T$ time steps. The signal is passed through a CNN to produce $I_{Ch}$  $\times$ $I_{Tsteps}$, where $I_{Ch}$ and $I_{Tsteps}$ is the number of input channels and timesteps for the BFM. We utilize categorical cross entropy loss for training the EEG adapter while finetuning the pretrained BFM. During the training process the CNN architecture utilizing temporal convolution inside the adapter learns to extract most relevant signal information. The temporal and spatial dependencies for each dataset are learnt by finetuning of the BFM without explicit mapping. Thus we have designed adapters for each dataset by varying the number of convolution layers.



\section{Experiments and Results}
In this section, we present the details of our experimental setups and results along with comparisons with contemporary methods.
\subsection{Experimental Setup}
We first filter EEG signal with a notch filter at 50Hz and a bandpass filter with a low cutoff of 0.1Hz and a high cutoff of 75Hz, as utilized by LaBraM. This preprocessing step is common across all datasets in our experiments. 
Our experiments are conducted on single 3080Ti GPU by Python 3.11.11, PyTorch 2.0.1 and CUDA11.8.
\subsection{Dataset Description}
\subsubsection{EEG-ImageNet}
This dataset consists of EEG signals capturing neural activity in response to visual stimuli from 40 object classes \cite{Palazzo2021}.
For this dataset, we have used the train, validation, and test split provided by the authors \cite{Palazzo2021}. We have a total of 7741, 1768 and 1790 samples in the train, validation, and test sets respectively.

To effectively utilize the existing BFM, we first convert the raw output which is the quantized value, \textbf{Q} of the signal to its microvolt, \textbf{V} representation using the formula:
\begin{equation}
\textbf{V} = \textbf{R} \times \textbf{Q} 
\end{equation}
Where resolution, \textbf{R} for each channel is provided in \cite{CVPR40resolution}. 

\subsubsection{BrainLat}
The dataset \cite{Prado2023} includes neurodegenerative diseases such as alzheimer's disease (AD), behavioral variant frontotemporal dementia (bvFTD), multiple sclerosis (MS), and parkinson's disease (PD), along with healthy controls (HCs). Due to inconsistencies in the MS and PD data across collection regions, these subsets were excluded from the experiments.
Typically, two setups are used for training deep learning models in medical diagnosis:
\begin{enumerate}
    \item Subject Dependent: Samples from each subject are present in all splits for training \cite{sdnour,sdkumar}, limiting real-world applicability. In real-time, subject information is not available prior to train the model.
    \item Subject Independent: A subject's data is in only one split \cite{subjectindependent_1,subjectindependent_2}, enhancing real-world applicability. This setup is more practical for medical diagnosis as subject information is not needed during training.
\end{enumerate}

In our experiments, we utilized the subject-independent setup. Each subject's EEG signals contain 10 minutes of continuous resting-state data. We applied a non-overlapping window of 128 timesteps to extract multiple time chunks, resulting in 108,278, 19,290, and 20,986 samples in the train, validation, and test sets, respectively.


.

\subsection{Results}

In this section we provide experimental results for visual brain decoding and resting state EEG datasets.
\subsubsection{EEG-ImageNet} 
\begin{figure}[h!]
\centering
\includegraphics[width=0.8\textwidth]{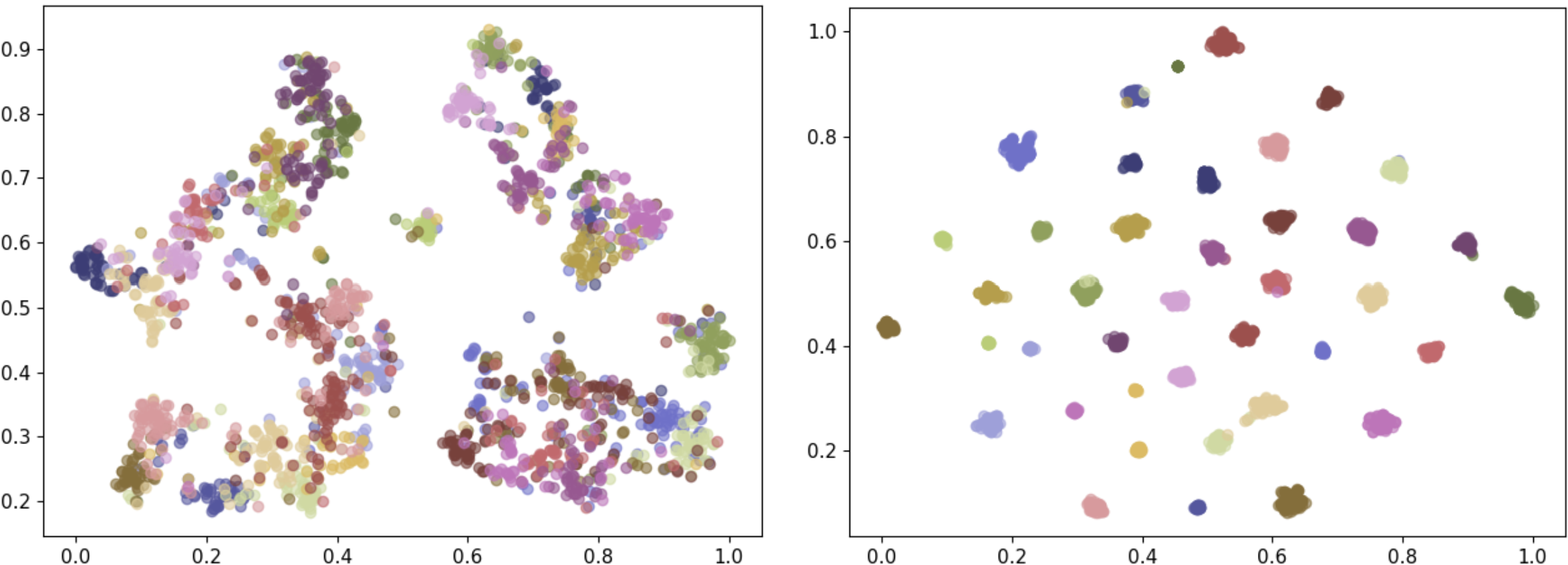}
\caption{On the left, the tSNE plot is for the manual channel selection without mixing data. On the right, the tSNE plot is for the automatic distillation using EAD. Each color in both plots denotes a different class. This provides evidence that using EAD, effective learning is happening for the data.} \label{manual_head_tsne}
\end{figure}

The current SOTA is BioLSTM which relies on LSTM based modelling. Using temporal convolutions in adapter in conjuction with BFM we achieve a higher accuracy as shown in Table \ref{pami2020_r1} alongwith tSNE \cite{VanderMaaten2008} plots in Figure \ref{manual_head_tsne} for manual channel selection and EAD. 
Also, to demonstrate the generalizability of our methodology via a zero shot setting, we followed the experimental setup used by \cite{Singh2024}, where six classes for each user are not present in any split during training phase. On top of the trained EAD as a feature extractor for the classes, our method outperforms on SVM \cite{Heart1998} classifier as shown in Table \ref{6unseen}.

\begin{table}[h!]
\centering
\caption{This table provides a comparison for the EEG-ImageNet dataset classification. We've compared our results with supervised as well as unsupervised methods.}\label{pami2020_r1}
\begin{tabular}{|l|l|l|l|l|}
\hline
Approach & Accuracy & Precision & Recall & F1 score\\
\hline
DML\cite{Jiang2019} & 97.70 & - & - & -\\
EEGLSTM\cite{Singh2024} & 98.30 & - & - & -\\
NeuroVision \cite{Khare2022}& 98.80 & - & - & -\\
BioLSTM\cite{Jiang2020}& 99.10 & - & - & -\\

\hline
BFM with channel selection & 67.64 & 68.36 & 67.64 & 67.53\\
BFM with channel mixing & 99.11 & 99.12 & 99.11 & 99.10\\
LaBraM with raw input& 98.84 &	98.88 &	98.84 &	98.84\\
EAD & {\bfseries 99.33} &{\bfseries 99.34} &{\bfseries 99.33} & {\bfseries 99.33}\\
\hline
\end{tabular}
\end{table}

\begin{table}[h!]
\centering
\caption{This provides the comparison for classification of 6 unseen classes for EEG-ImageNet}\label{6unseen}
\begin{tabular}{|l|l|l|l|}
\hline
Approach & SVM & KNN & K-means\\
\hline
Cogni-Net\cite{Mukherjee2019}& 78.00 &	72.50& -\\
EEGLSTM\cite{Singh2024}& 93.00 & {\bfseries 86.00}&	{\bfseries 62.50}\\
EAD & {\bfseries 98.21} & 75.11 & 39.46\\
\hline
\end{tabular}
\end{table}

\subsubsection{BrainLat}
\begin{figure}[h!]
\centering
\includegraphics[width=0.5\textwidth]{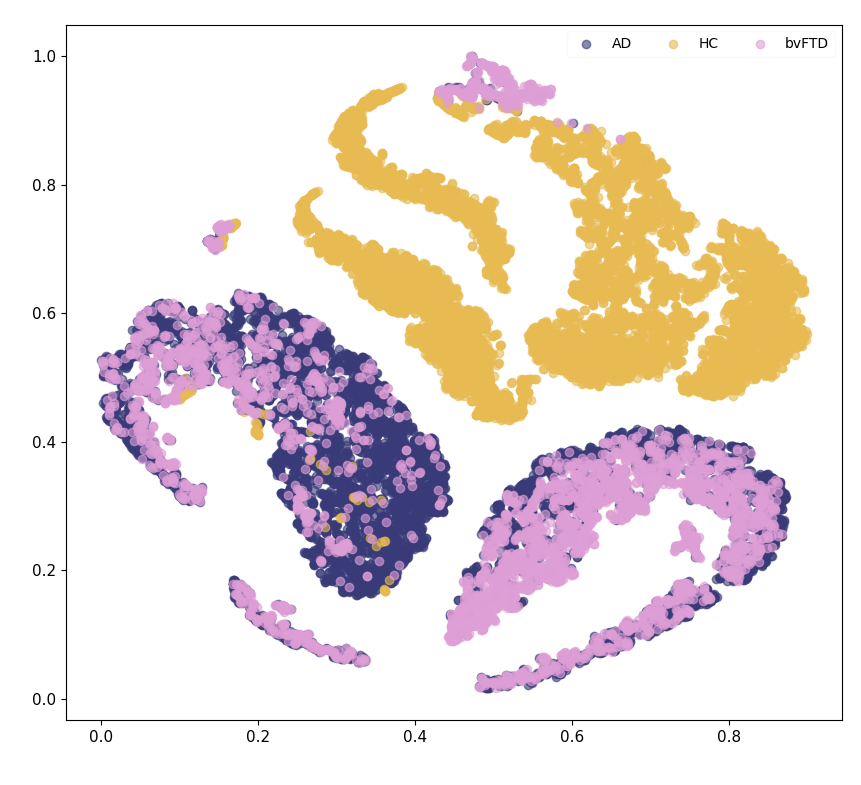}
\caption{tSNE plot for BrainLat dataset. } \label{brainlat_tsne}
\end{figure}

For this dataset, we present the subject independent results for the first time, detailing the EAD outcomes at both the sample and subject levels in Table \ref{brainlat_r}. Additionally, the tSNE plot for the test set is illustrated in Figure \ref{brainlat_tsne}. It is evident from clusters that the model performs extremely good at differentiating between HC and disease groups but gets confused between AD and bvFTD.
\begin{table}[h!]
\centering
\caption{Results for sample level 
 and subject level classification.}\label{brainlat_r}
\begin{tabular}{|l|l|l|l|l|}
\hline
Approach & Accuracy & Precision & Recall & F1 score\\
\hline
EAD Sample Level Classification& 80.79 & 81.23 & 80.79 & 80.80\\
EAD Subject Level&	92.31&94.23&92.31&92.43\\

\hline
\end{tabular}
\end{table}

\section{Conclusion}
In this work we show that using EAD configuration, with LaBraM as BFM, we are able to effectively learn EEG signal representation for two very diverse tasks involving a resting state and visual brain decoding task. This paves the path for a generalized model for multiple avenues of EEG signal classification and conditional image generation. A future direction will involve achieving further montage invariance involving low density as well as high density EEG datasets.

\begin{credits}
\subsubsection{\ackname} The work has been partly supported by Science and Engineering Research Board, Government of India.
\end{credits}

\end{document}